\documentclass[letterpaper, 10 pt, conference]{ieeeconf}  
\usepackage{multirow}
\usepackage{graphicx}
\usepackage{booktabs}
\usepackage{subfigure}
\usepackage{caption}

\usepackage{algorithm, algorithmic}
\usepackage{amsmath,amssymb}
\usepackage{color}
\usepackage{url}
\usepackage[misc]{ifsym} 
\usepackage{lineno,hyperref}
\usepackage{biblatex}

\usepackage{soul,color,xcolor}

\IEEEoverridecommandlockouts                              

\title{\LARGE \bf
One Training for Multiple Deployments: Polar-based Adaptive BEV Perception for Autonomous Driving
}

\author{Huitong~Yang$^{1}$, Xuyang~Bai$^{2}$, Xinge~Zhu$^{3\star}$, and Yuexin~Ma$^{1\star}$
\thanks{$^{\star}$ Corresponding author}
\thanks{$^{1}$Huitong~Yang and Yuexin~Ma is with School of Information Science and Technology, ShanghaiTech University, Shanghai 201210, China. {\tt\small mayuexin@shanghaitech.edu.cn}}%
\thanks{$^{2}$Xuyang Bai is with Department of Computer Science and Engineering, Hong Kong University of Science and Technology. {\tt\small xbaiad@connect.ust.hk}}%
\thanks{$^{3}$Xinge Zhu is with Department of Information Engineering, the Chinese University of Hong Kong.  {\tt\small zx018@ie.cuhk.edu.hk}}%
}

\begin{document}

\maketitle
\thispagestyle{empty}
\pagestyle{empty}

\begin{abstract}

Current on-board chips usually have different computing power, which means multiple training processes are needed for adapting the same learning-based algorithm to different chips, costing huge computing resources. The situation becomes even worse for 3D perception methods with large models. Previous vision-centric 3D perception approaches are trained with regular grid-represented feature maps of fixed resolutions, which is not applicable to adapt to other grid scales, limiting wider deployment. In this paper, we leverage the Polar representation when constructing the BEV feature map from images in order to achieve the goal of training once for multiple deployments. Specifically, the feature along rays in Polar space can be easily adaptively sampled and projected to the feature in Cartesian space with arbitrary resolutions. To further improve the adaptation capability, we make multi-scale contextual information interact with each other to enhance the feature representation. Experiments on a large-scale autonomous driving dataset show that our method outperforms others as for the good property of one training for multiple deployments. 

\end{abstract}

\section{INTRODUCTION}

Accurate 3D perception is critical for autonomous vehicles to understand the complexities surrounding scenes and make safe and effective driving decisions. There are tremendous research works contributing to this task, among which the vision-centric methods become more and more popular due to the low computation cost. Although such learning-based approaches have achieved promising performance, they are usually trained with a predefined feature map resolution, which can only be integrated into limited on-board chips with corresponding computing resources and are not applicable for multiple deployments under arbitrary computational budgets. Retraining a model with different resolutions will cost extra huge computing resources. Moreover, perception is not the only task conducted on the chip, other tasks also occupy computing resources. Dynamic resource allocation can improve the service efficiency of the chip, but it demands that the perception algorithm possesses dynamic adjustment capability as well. Thus, a vision-centric method that can adaptively adjust the computation requirements and be deployed to chips with various computing performances becomes extremely significant.

\begin{figure}[htbp]
	\centering
	\includegraphics[width=1.0\columnwidth]{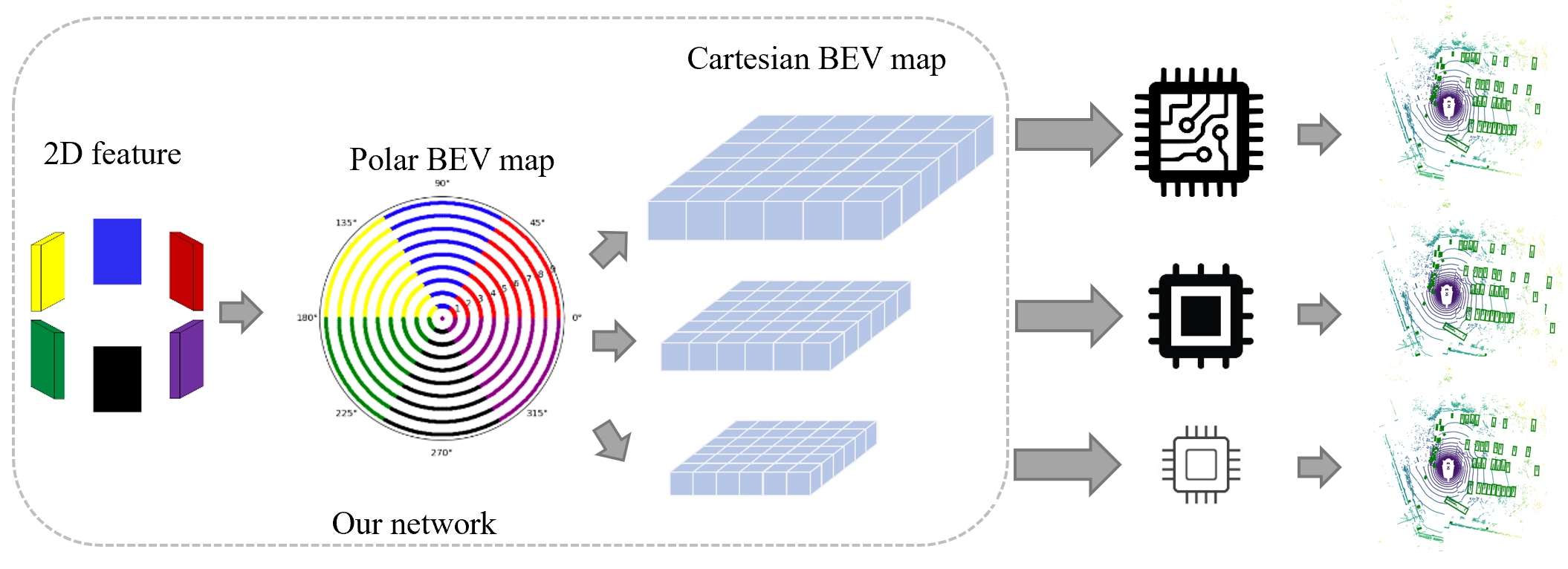}
	\caption{Benefiting from the grid-insensitive sampling nature of the Polar-based feature map representation, our network can supply various Cartesian BEV grid resolutions for 3D perception according to the actual computing budget during deploying.}
	\label{teaser}
\vspace{-2ex}
\end{figure}

Current mainstream vision-centric 3D perception methods usually project the perspective-view (PV) image feature map to the bird's eye view (BEV) feature map and then attach detection or segmentation head for specific 3D perception tasks. However, such methods construct regular-grid represented BEV feature maps in the Cartesian coordinate system and assign fixed physical scale correspondences for the grid during training, making them hard to be adapted to other resolutions. Directly downsampling or upsampling the BEV feature map to satisfy various computing requirements will result in dramatic performance drop for 3D perception. 
Actually, compared with regular grid representation, Polar representation on BEV is more reasonable since each Polar ray can physically correspond to one column of the image pixels~\cite{image2map,PolarBEV} according to foreshortening effect of camera imaging. More importantly, the feature along each ray is grid-insensitive, and can be flexibly projected to multiple Cartesian feature maps with different resolutions, as shown in Figure.~\ref{teaser}. That is to say, even though the network is trained under the condition that the Polar feature is transferred to $128\times128$ Cartesian feature map, the pre-trained Polar feature can easily adapt to new Cartesian feature maps with the resolution of $256\times256$ or $64\times64$ to fit diverse computing budget allocations. To our knowledge, we are the first to exploit this property of Polar-based BEV representation to achieve the goal of one training for multiple deployments.

In this paper, based on Polar BEV feature representation, we propose a novel vision-centric 3D detection method for autonomous driving, which can be flexibly deployed multiple times to adapt to diverse computing budgets with only one training procedure. The whole pipeline contains three main modules, as Figure.~\ref{fig:network} shows, including the column-wise PV-to-BEV transformer module, the grid-insensitive Polar-to-Cartesian sampling module, and the multi-scale feature interaction module. The first module projects the PV image feature to BEV Polar feature by multi-head attention and depth distribution estimation. 
For the second module, the feature along rays in the Polar coordinate system can be adaptively sampled and projected to the feature with different grid sizes in Cartesian space. Therefore, when the computing budget becomes limited, our network can naturally transfer the Polar feature to small Cartesian feature maps to reduce the occupancy of computing resources and keep high-quality performance at the same time.
To further improve the adaptation capability, we design the multi-scale feature interaction module to enhance the feature representation through transformer mechanism. At the end, we attach one detection head for the 3D object localization. We evaluate our method on a large-scale autonomous-driving dataset, nuScenes~\cite{nuscenes}, and achieve the state-of-the-art performance under the condition of one training for multiple deployments. Ablation study is also conducted to verify the effectiveness of important modules of our network.

In summary, our contributions are as follows,

\begin{itemize}
\item We propose a novel Polar-based BEV perception method, which can be adapted to various computing budgets for multiple deployments once trained.
\item We make use of the information interaction among multi-scale BEV features to enhance the feature representation for better adaptation.
\item Our method achieves state-of-the-art generalization capability of inferencing on feature maps with unseen resolutions for 3D detection on large-scale autonomous driving dataset.
\end{itemize}

\section{RELATED WORK}
With the popularization of low-cost autonomous driving technology, vision-centric BEV perception algorithms have gradually attracted attention. Our work focuses on this promising technique and further proposes an adaptive deployment solution, which is urgent and significant for real applications. We introduce the related work from two aspects, including Cartesian-based BEV perception methods and polar-based ones.

\subsection{Cartesian-based BEV Perception}
 Current BEV perception works~\cite{BEVSurvey} can mainly be divided into homograph-based, depth-based, MLP-based methods, and transformer-based four kinds of methods. Homograph-based methods~\cite{GA,TrafCam,VPOE,BEVstitch,3D-LaneNet,OGM,DSM,cam2BEV} exploit the Inverse Perspective Mapping (IPM)~\cite{IPM} to establish the geometric transformation between PV and BEV view so that many downstream vision tasks can be explored, such as 3D object detection, semantic map construction, and motion planning. In addition, depth-based methods~\cite{OFT,BEVDet,beverse,CADDN,panopticseg,StretchBEV} estimate depth distribution explicitly or implicitly and use it as the bridge to generate BEV feature from multi-view image feature. LSS~\cite{LSS,BEVDet} builds the camera frustum by pixel-level depth distribution and projects it to the world space. To obtain the more accurate depth estimation, BEVDepth~\cite{BEVDepth} leverages the depth value of sparse Lidar points to provide explicit depth supervision, which significantly improves the accuracy of 3D detection. Moreover, some methods~\cite{FishingNet,PYVA,HFT,VPN,Hdmapnet,PON,STA-ST} utilize MLP as a mapping function to transform views. For transformer-based methods~\cite{petr,petrv2,ora3d,bevsegformer,panoticsegformer}, they usually design a set of BEV queries using positional encoding, then performs the view transformation through cross attention between BEV queries and the image features. DETR3D~\cite{DETR3D} first applies sparse 3D learn-able reference points as no-explicit BEV queries to link the 2D feature from multi-view cameras. BEVFormer~\cite{bevformer} exploits the transformer to unify the spatial and temporal information to learn the dense BEV feature without depth estimation and enable multi-task learning. Although above algorithms exhibit competitive performance, their feature maps are all designed in Cartesian coordinate system, where regular grid division is defined with fixed scale correspondence to the physical world. If we use such networks trained under one scale to infer other scales of output, the performance will dramatically drop.

\subsection{Polar-based BEV Perception}
Polar coordinate system is another choice to represent the 3D space. It first appears in LiDAR-based 3D perception~\cite{SSN,Cong_2022_CVPR} approaches. PolarNet~\cite{PolarNet} projects the point cloud to BEV representation in Polar coordinate, and controls the points with learnable grid cells, bringing significant effects to the long-tailed problem of spatial point cloud distribution. Cylinder3D~\cite{Cylinder3D} proposes dynamical 3D voxelization in the cylinder coordinates to adapt the varying-density property of outdoor Lidar points. Actually, polar representation is also applicable in the camera-based BEV perception task due to the foreshortening effect of camera imaging. Moreover, the Polar-based BEV representation~\cite{polarformer} is more consistent to represent the whole scenes captured by the ego vehicle's surrounding cameras. Following the ray tracing principle, Ego3RT~\cite{Ego3RT} treats the dense polar BEV grids as human eyes and uses camera parameters to construct a 3D-to-2D reverse projection for multi-task learning. PolarBEV~\cite{PolarBEV} establishes an uneven BEV grid representation, considering the different object scales according to the distance to the cameras by the angle-specific and radius-specific embedding. Our method also projects the image columns to polar rays with the transformer. However, we are the first to focus on the BEV grid insensitive sampling nature of the polar map and leverage the property to achieve multi-deployment adaptation according to the real-time computational budget and inference speed in actual applications.

\section{Method}
\begin{figure*}
    \centering
	\includegraphics[width=1.0\textwidth]{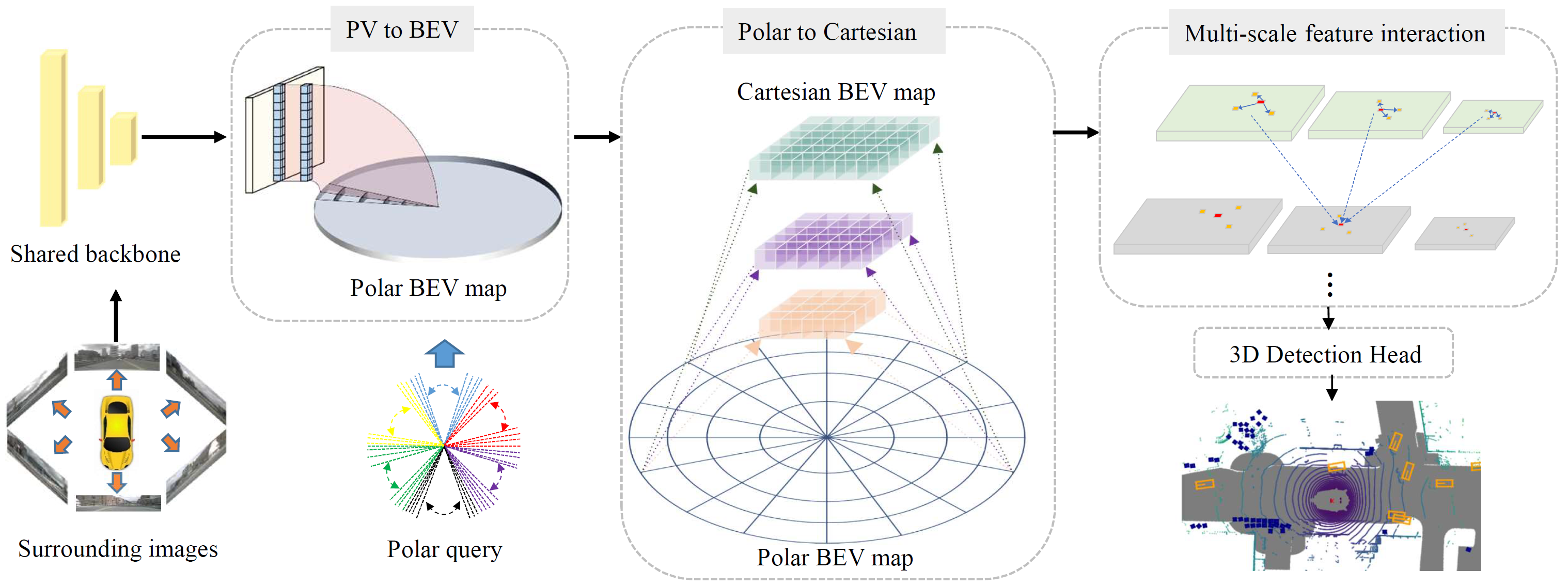} 
\caption{Overall pipeline of our method. We exploit standard 2D backbone to extract multi-view features. Then, a Column-wise PV-to-BEV Transformer module projects the columns from image feature map to corresponding Polar rays using a cross-attention mechanism. Next, we propose a grid-insensitive Polar-to-Cartesian sampling strategy to transform the Polar BEV map to various granularities of the Cartesian BEV grid, which can be adjusted during deployment according to the computation budgets in the actual autonomous scenes. We additionally propose a multi-scale BEV feature interaction encoder to correlate contextual information across multi-scale Cartesian BEV map. Finally, a detection head is attached to predict objects' 3D bounding boxes and categories.}
\vspace{-2ex}
\label{fig:network}
\end{figure*}

We propose a novel vision-centric Polar-based 3D object detector, which is trained once and can be deployed
with various granularity of BEV feature maps based on the computation budget.
As shown in Fig~\ref{fig:network}, our framework takes multi-view RGB images as input and produces 3D bounding boxes and their category as output. Following~\cite{BEVDet}, the surround-view images 
are first encoded by a single shared backbone to obtain multi-scale 2D representations 
$\{{{F}_{i}}\}_{i=1}^{{scale}}$ with size of ${\frac{H}{{{2}^{i+1}}}\times \frac{W}{{{2}^{i+1}}}\times C}$
and then unifed into a single resolution 2D feature map using FPNLSS.
To conduct 3D perception tasks from images, the mainstream methods usually start with projecting the PV feature map onto BEV space, which can be directly deployed by many downstream real-world applications such as behavior prediction, motion planning, etc. 
However, since they usually adopt a regular
grid-based Cartesian BEV representation with a predefined grid size during training, it is difficult to adapt to different scales during inference, limiting their applications in self-driving scenarios that are having different computational budgets from the training process.
To mitigate this issue and design a detector that can be flexibly deployed under different computational budgets, we adopt the Polar coordinate representation, which represents the physical world more naturally according to the image-forming principle. 
More importantly, the feature along each ray is grid-insensitive, and can be utilized to project into multiple Cartesian BEV feature maps with different resolutions.
In this way, our method can avoid multiple times of training for deployments under different computational budgets, saving lots of time and resources, which is essential to applications on limited on-board chips. To achieve this goal, we propose three novel modules, including the column-wise PV-to-BEV transformer module, the grid-insensitive Polar-to-Cartesian sampling module, and the multi-scale feature interaction module, which will be illustrated in the following.

\subsection{Column-wise PV-to-BEV Transformer Module}
One key step in vision-centric perception algorithms is to perform view transformation from PV to BEV, as BEV is a natural representation of the world and is fusion-friendly. The mainstream methods can be divided into two categories: geometry-based transformation and network-based transformation~\cite{BEVSurvey}. 
The former fully utilizes the physical principles of the camera to transfer the view in an interpretable manner, but might suffer from inaccurate depth prediction. The latter performs the view transformation in a data-driven way without explicitly leveraging the camera projection model. 
In our work, we build a network-based view transformation module by extending the single-image view transformer \cite{image2map} to multiview scenario. As shown in Fig.~\ref{fig:network}, we first assume the 1-1 relationship between vertical scanlines in the image and the rays on the BEV plane starting from the camera center. Then the view transformation can be formulated as a set of 1D sequence-to-sequence translation problems and modeled by a transformer module. Such column-wise transformer module avoids the dense attention between 2D image feature maps and BEV queries, and instead only relies on 1D sequence-to-sequence translation, leading to a memory-friendly and data-efficient architecture. 

\begin{figure}[t]
	\centering
	\includegraphics[width=1\columnwidth]{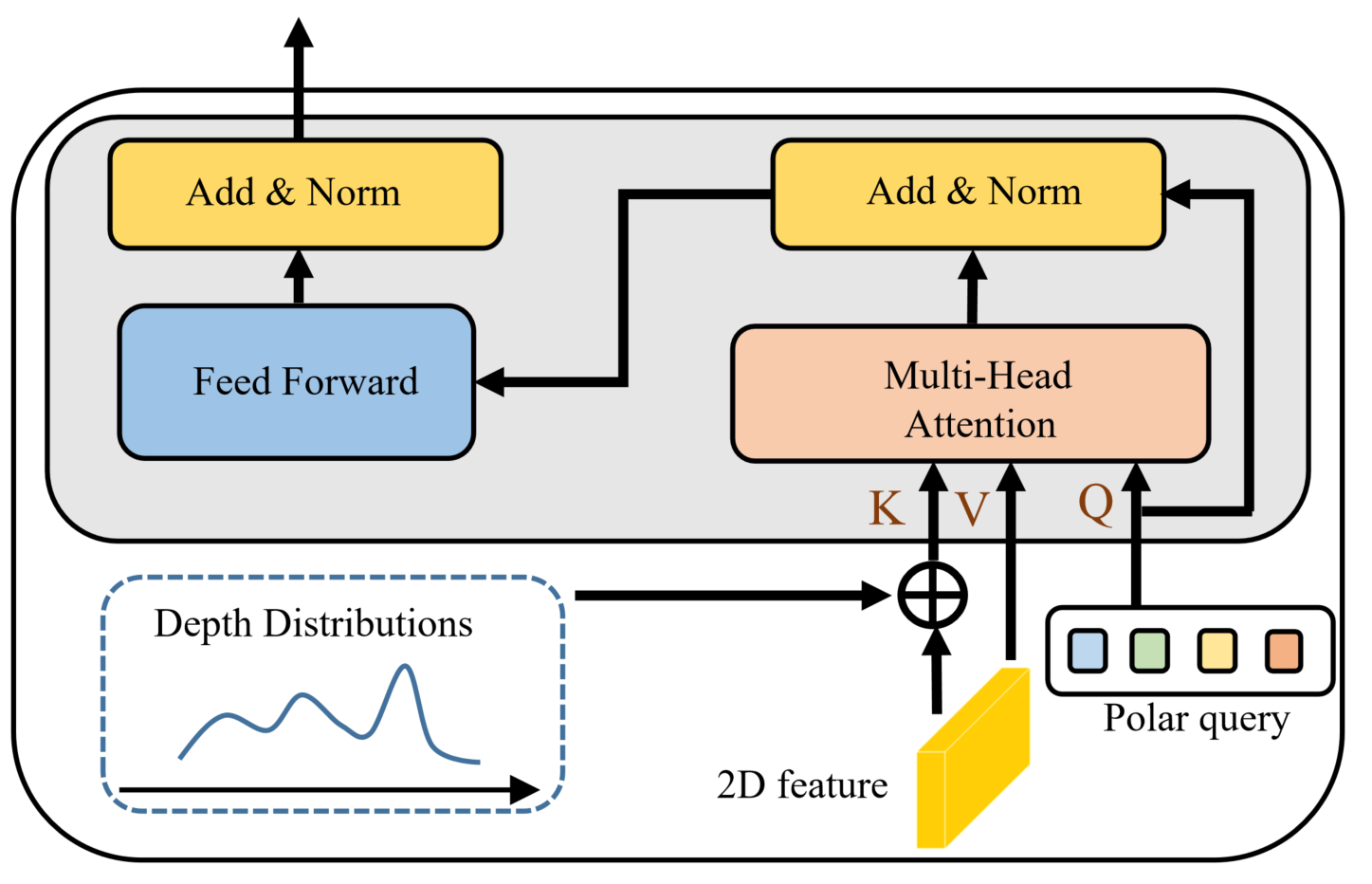}
	\caption{Detailed network of the column-wise PV-to-BEV transformer module. }
	\label{cross-atten}
 \vspace{-3ex}
\end{figure}

Specifically, we first design a set of positional embedding~(called Polar query) initialized on the polarized BEV plane. Next, based on the mapping relationship between each image column and Polar ray established by the camera geometry, we translate the image features from the image column to its associated ray by performing cross attention between image features on the column and the Polar queries on the ray. This process can be viewed as a learnable assignment of semantic PV features to their positional slots along a Polar ray in the BEV plane. Different from the single image case in \cite{image2map}, multiple surround images usually overlap with each other, where the multiview observations of the overlap region is critical for many detection targets, such as the large objects being truncated in a single image. Thus to enhance the information on overlap regions, we count the number of columns projected to the same ray and adaptively assign higher attention weights, so that our column-wise cross-attention module can better handle the information from overlapping regions.

To further facilitate the geometric reasoning ability of transformer module, we inject the depth information into the view transformation process, as shown in Fig~\ref{cross-atten}. We first predict the depth distribution of each image pixel following \cite{BEVDepth} and then use an MLP network to embed the depth distribution into positional embedding, which will be added with the 2D features to provide the depth guidance for PV-to-BEV transformation. Although our column-wise PV-to-BEV transformation module does not necessarily need per-pixel depth for view transformation, we still empirically find the depth information to be important for geometric reasoning of transformers.

\subsection{Grid-insensitive Polar-to-Cartesian Sampling Module}
To enable the deploy-time adjustment of perception granularity based on the actual computation budget, 
we propose a grid-insensitive sampling module so that our network can generalize to different BEV feature resolutions. Specifically, we first generate a Cartesian meshgrid $M \in R^{H\times W\times 2}={(w_i, h_i)}$ within the multi-view perception range $[-51.2m,51.2m]$ with corresponding resolutions, where $H\times W$ is the spatial resolution of BEV feature map. The meshgrid will be leveraged during Polar-to-Cartesian feature transformation by sampling the Polar map to get the Cartesian BEV feature map.
The conversion between the Polar coordinate system and Cartesian coordinate system is calculated as follow,
\begin{equation}
\begin{aligned}
    {{\phi }^{r}}=\arctan (\frac{{{w}_{i}}}{{{h}_{i}}}),
    {{\sigma }^{r}}=\sqrt{({{w}_{i}}^{2}+{{h}_{i}}^{2})},
\end{aligned}
\end{equation}
where $({\phi }^{r},{\sigma }^{r})$ is the coordinate of Polar map. 
To cover the perception range of $[-51.2m,51.2m]$, we set the range of Polar coordinate system to ${\sigma^{r}}\in [0m,72m], \phi^r \in [0, 2\pi]$
and normalize the Polar-to-Cartesian transformation. Finally, we stack the normalized Polar coordinate to get the Polar-to-Cartesian BEV meshgrid:

\begin{equation}
\begin{aligned}
{{\phi }^{r}}=({{\phi }^{r}}-P_{\min }^{r})/(P_{\max }^{r}-P_{\min }^{r}), \\
{{\sigma }^{r}}=({{\sigma }^{r}}-\sigma _{\min }^{r})/(\sigma _{\max }^{r}-\sigma _{\min }^{r}), \\ 
{{P}=\text{Stack}([{{\phi }^{r}},{{\sigma }^{r}})]\in {{R}^{H\times W\times 2}}.
}
\end{aligned}
\end{equation}

Then our grid-insensitive sampler can transform the Polar BEV map to Cartesian BEV grid with different resolutions through the bi-linear interpolation operation. The multi-resolution BEV feature map derived from the grid-insensitive sampler enables the potential of adapting our single network to various granularities of feature maps during inference to facilitate the deployment of different computational budgets.

\subsection{Multi-scale Feature Interaction Module}
To capture multi-scale contextual information and improve the representation power of multiple BEV feature maps with different resolutions,
we propose a Multi-scale BEV Interaction Encoder~(MBIE) to extract different granularity of contextual information from sampled Cartesian BEV maps. 
Following \cite{polarformer}, we build the MBIE module by stacking multiple deformable attention layers to fuse the information from different scales.
Formally, the deformable attention operator is defined as:

\begin{equation}
\begin{aligned}
    &\text{MSBEVDeformAttn}({{b}_{q}},\overset{\wedge }{\mathop{{{r}_{q}}}}\,,\{{{b}^{s}}\}_{s=1}^{S})= \\ &\sum\limits_{h}^{H}{{{W}_{h}}[\sum\limits_{s=1}^{S}{\sum\limits_{r=1}^{R}{{{A}_{hsqr}}{{W}_{h}}^{'}}}{{b}^{s}}({{\sigma }_{s}}(\overset{\wedge }{\mathop{{{r}_{q}}}}\,)+\Delta {{f}_{hsqr}})]},
\end{aligned}
\end{equation}
where ${{q}}$ is a query element in the BEV query feature ${{b}_{q}}$, $\overset{\Lambda }{\mathop{{{r}_{q}}}}\,\in {{[0,1]}^{2}}$ represents the normalized coordinates of reference points, ${{b}^{s}}$ is the multi-scale BEV features, ${{\sigma }_{s}}(*)$ is the rescaling operator that converting ${{r}_{q}}$ to corresponding scale and  $\Delta {{f}_{hsqr}}$ is the offset that helps the query feature interacting with the neighbour contextual BEV information from all levels. MBIE performs information exchange among neighboring pixels and across multiple Cartesian BEV feature maps, which enhances the multi-scale context for each feature map, 
and serves as a crucial step for the generalization to different BEV feature resolutions.


Followed by the MBIE module, a deformable convolution based FPN module~\cite{Huang2021FaPNFP} is leveraged to fuse the BEV feature map with different resolutions into the target resolution. Since we can manually control the down-sampling or up-sampling factors in FPN, it is convenient to adjust the resolution of output feature during inference. More details about the inference strategy will be illustrated in Sec.~\ref{subsec:imp1}. 

Given the dense BEV feature map generated from the previous stages, we adopt the first stage of CenterPoint as our detection head due to its simplicity and popularity. Following \cite{centerpoint}, we adopt the heatmap loss for classification supervision and the L1 regression loss for bounding box supervision. 
\begin{table*}[htbp]
  \centering
  \caption{In this table, we show the result on different output resolutions with fixed training resolution, which also meets our motivation, \emph{i.e.}, one training for multiple deployments. - means the value is not available for the method.}
    \scalebox{0.82}{\begin{tabular}{c|c|c|c|c|c|c|c|c|c|c}
    \hline
    Method & BEV Grid Size & Output Resolution & Training Head & NDS$\uparrow$   & mAP$\uparrow$   & mATE$\downarrow$  & mASE$\downarrow$  & mAOE$\downarrow$  & mAVE$\downarrow$  & mAAE$\downarrow$ \\
    \hline
    \multirow{3}[2]{*}{BEVDet} & 0.4 Meter & 128×128 & \multirow{3}[2]{*}{128×128} & 0.392 & 0.312 & 0.691 & \textbf{0.272} & 0.523 & 0.909 & 0.247 \\
          & 0.256 Meter & 200×200 &       & 0.324 & 0.244 & 0.888 & 0.308 & 0.606 & 0.958 & 0.211 \\
          & 0.2 Meter & 256×256 &       & 0.275 & 0.201 & 1.004 & 0.328 & 0.702 & 1.010 & 0.221 \\
    \hline
    \multirow{3}[2]{*}{BEVerse} & 0.4 Meter & 128×128 & \multirow{3}[2]{*}{128×128} &   -   & 0.321 &   0.682   &  0.278   &  \textbf{0.464}   &  \textbf{0.327}   &  \textbf{0.189} \\
          & 0.256 Meter & 200×200 &       &   -   & 0.275 &   0.804   &  0.318   &  0.543   &  \textbf{0.517}   &  \textbf{0.195} \\
          & 0.2 Meter & 256×256 &       &   -   & 0.230 &  0.898   &  0.337   &  0.615   &  \textbf{0.662}   &  \textbf{0.194} \\
    \hline
    \multirow{3}[2]{*}{Ours} & 0.4 Meter & 128×128 & \multirow{3}[2]{*}{128×128} & \textbf{0.398} & \textbf{0.321} & \textbf{0.669} & 0.275 & 0.494 & 0.956 & 0.231 \\
          & 0.256 Meter & 200×200 &       & \textbf{0.378} & \textbf{0.305} & \textbf{0.710} & \textbf{0.280} & \textbf{0.519} & 1.024 & 0.241 \\
          & 0.2 Meter & 256×256 &       & \textbf{0.346} & \textbf{0.278} & \textbf{0.789} & \textbf{0.300} & \textbf{0.568} & 0.991 & 0.252 \\
    \hline
    \end{tabular}}%
  \label{tab:seen}%
\end{table*}%

\begin{table*}[htbp]
  \centering
  \caption{Results of more challenging setting. It can be found that these output resolutions are totally unseen and flexible.}
    \scalebox{0.82}{\begin{tabular}{c|c|c|c|c|c|c|c|c|c|c}
    \hline
    Method & BEV Grid Size & Output Resolution & Training Head & NDS$\uparrow$   & mAP$\uparrow$   & mATE$\downarrow$  & mASE$\downarrow$  & mAOE$\downarrow$  & mAVE$\downarrow$  & mAAE$\downarrow$ \\
    \hline
    \multirow{2}[2]{*}{BEVDet} & 0.8 Meter & 64×64 & \multirow{2}[2]{*}{128×128} & 0.181 & 0.092 & 1.078 & 0.505 & 0.781 & 0.877 & 0.486 \\
          & 0.53 Meter & 96×96 &       & 0.336 & 0.231 & 0.860 & 0.301 & 0.575 & 0.842 & 0.217 \\ & 0.4 Meter & 128×128 & & 0.392 & 0.312 & 0.691 & \textbf{0.272} & 0.523 & 0.909 & 0.247\\
    \hline
    \multirow{2}[2]{*}{BEVerse} & 0.8 Meter & 64×64 & \multirow{2}[2]{*}{128×128} &   -   & 0.107 &  \textbf{1.054}   &  0.365   &  0.641   &  \textbf{0.510}   &  \textbf{0.251} \\
          & 0.53 Meter & 96×96 &       &   -   & 0.254 &  0.778   &  0.296   &  0.509   &  \textbf{0.385}   &  \textbf{0.187} \\ & 0.4 Meter & 128×128  & &   -   & 0.321 &   0.682   &  0.278   &  \textbf{0.464}   &  \textbf{0.327}   &  \textbf{0.189} \\
    \hline
    \multirow{2}[2]{*}{Ours} & 0.8 Meter & 64×64 & \multirow{2}[2]{*}{128×128} & \textbf{0.290}  & \textbf{0.143}      & 1.057   & \textbf{0.288}   & \textbf{0.589}   & 0.713  &  0.217 \\
          & 0.53 Meter & 96×96 &       & \textbf{0.370} & \textbf{0.280} & \textbf{0.749} & \textbf{0.279} & \textbf{0.504} & 0.951 & 0.223 \\
          & 0.4 Meter & 128×128 &  & \textbf{0.398} & \textbf{0.321} & \textbf{0.669} & 0.275 & 0.494 & 0.956 & 0.231\\
    \hline
    \end{tabular}}%
  \label{tab:unseen}%
\end{table*}%

\section{Experiment}
In this section, we first introduce the dataset, evaluation metrics, and implementation details. Then, we give detailed comparison of generalization capability with different baseline methods. Furthermore, we evaluate our method with various methods on nuScenes dataset to show its effectiveness and extensive ablation studies are conducted to validate each proposed component.

\subsection{Datasets \& Metrics}
We evaluate our approach on a large-scale dataset, nuScenes~\cite{nuscenes}, which contains 1000 scenes with durations of 20 seconds, and each frame contains 6 surrounding cameras with the resolution of $1600\times 900$.  For 3D object detection, we mainly evaluate the performance by the following metrics: mean Average Precision (mAP), Average Translation Error (ATE), Average Scale Error
(ASE), Average Orientation Error (AOE), Average Velocity Error (AVE), Average Attribute Error (AAE),
and NuScenes Detection Score (NDS) which is computed as (${\text{NDS}=\frac{1}{10}[5*\text{mAP}+\sum\nolimits_{mTP\in \mathbf{TP}}{(1-\min (1,\text{mTP}))}]}$).

\begin{table}[htbp]
  \centering
  \caption{The ablations on our proposed modules. CPBT represents the Column-wise PV-to-BEV Transformer Module. MBIE is the Multi-scale Feature Interaction Module.}
    \begin{tabular}{c|c|cc}
    \hline
    CPBT  & MBIE  & NDS$\uparrow$   & mAP$\uparrow$\\
    \hline
          &       & 0.384 & 0.294 \\
    \checkmark     &       & 0.391 & 0.307 \\
    \checkmark     & \checkmark     & \textbf{0.398} & \textbf{0.321} \\
    \hline
    \end{tabular}%
  \label{tab:ablation}%
\end{table}%

\begin{figure}[htbp]
\centering

\hspace{-1cm}
{
\begin{minipage}{0.01\textwidth} 
\centering
\rotatebox{90}{\footnotesize{$64\times64$}}
\end{minipage}
}\hspace{-0.1cm}
{
\begin{minipage}{0.12\textwidth} 
\centering
\includegraphics[height=2.5cm,width=2.5cm]{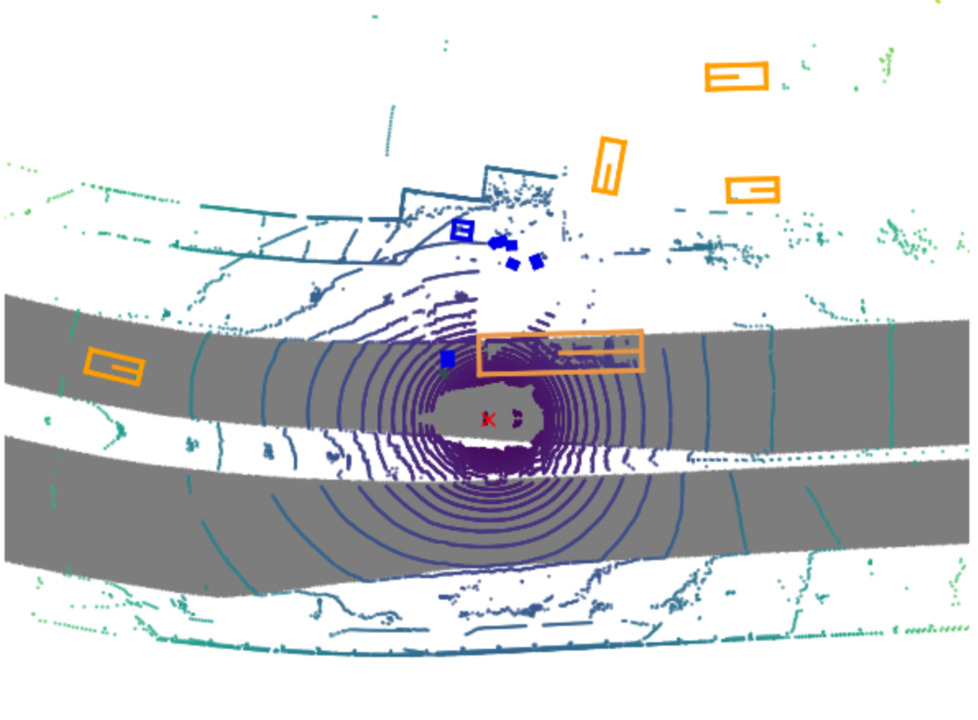}
\end{minipage}
}\hspace{0.27cm}
{
\begin{minipage}{0.12\textwidth} 
\centering
\includegraphics[height=2.5cm,width=2.5cm]{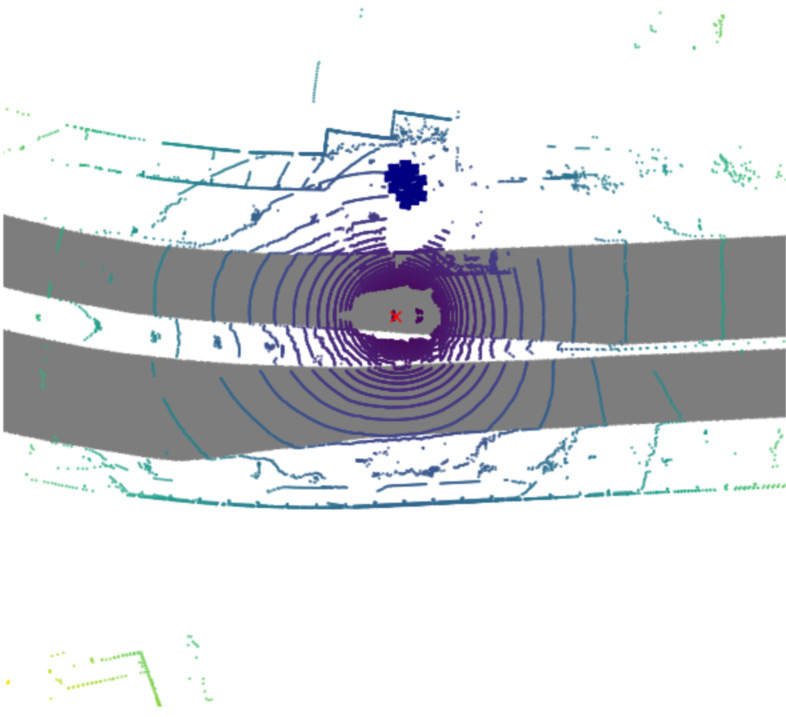}
\end{minipage}
}\hspace{0.27cm}
{
\begin{minipage}{0.12\textwidth} 
\centering
\includegraphics[height=2.5cm,width=2.5cm]{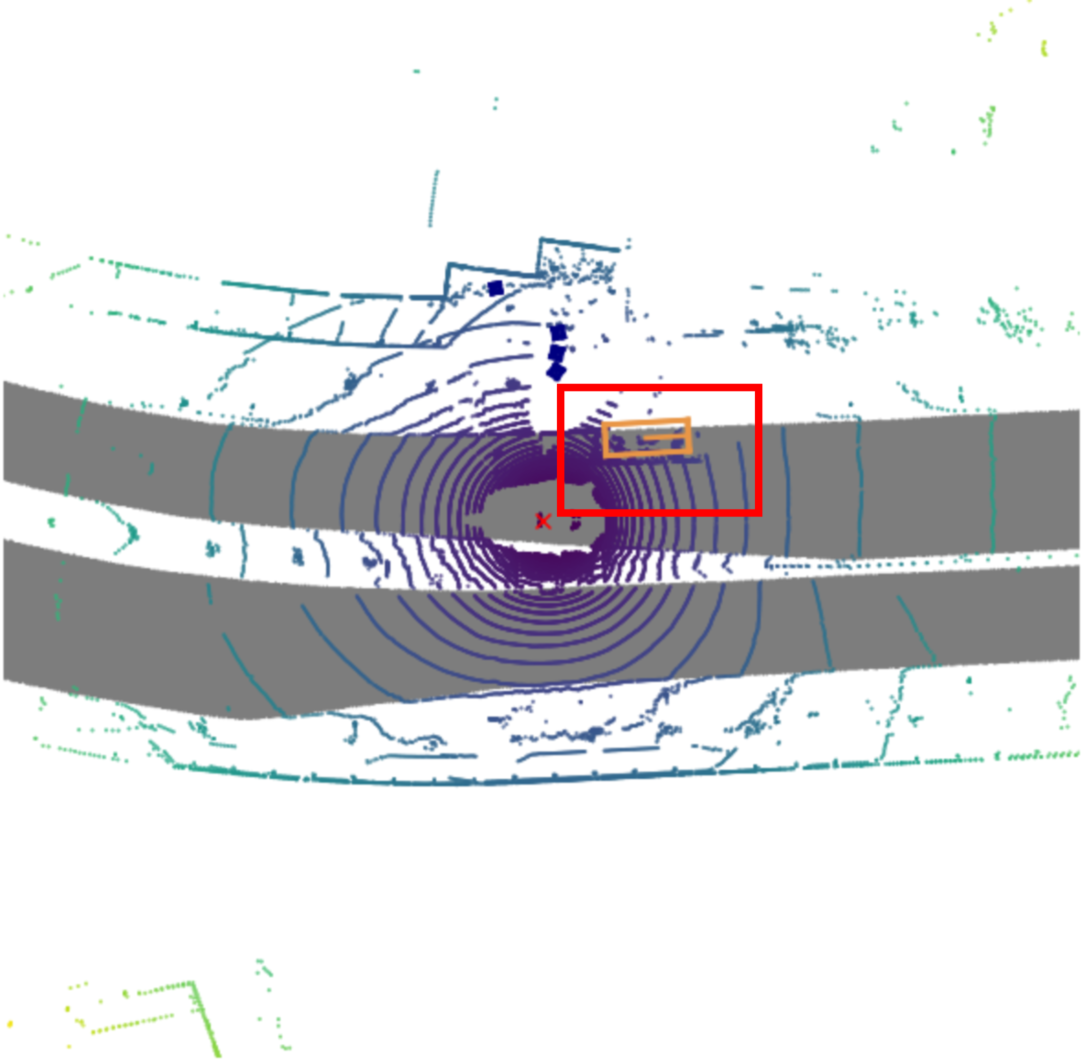}
\end{minipage}
}

\hspace{-1cm}
{
\begin{minipage}{0.01\textwidth} 
\centering
\rotatebox{90}{\footnotesize{$96\times96$}}
\end{minipage}
}\hspace{-0.1cm}
{
\begin{minipage}{0.12\textwidth} 
\centering
\includegraphics[height=2.5cm,width=2.5cm]{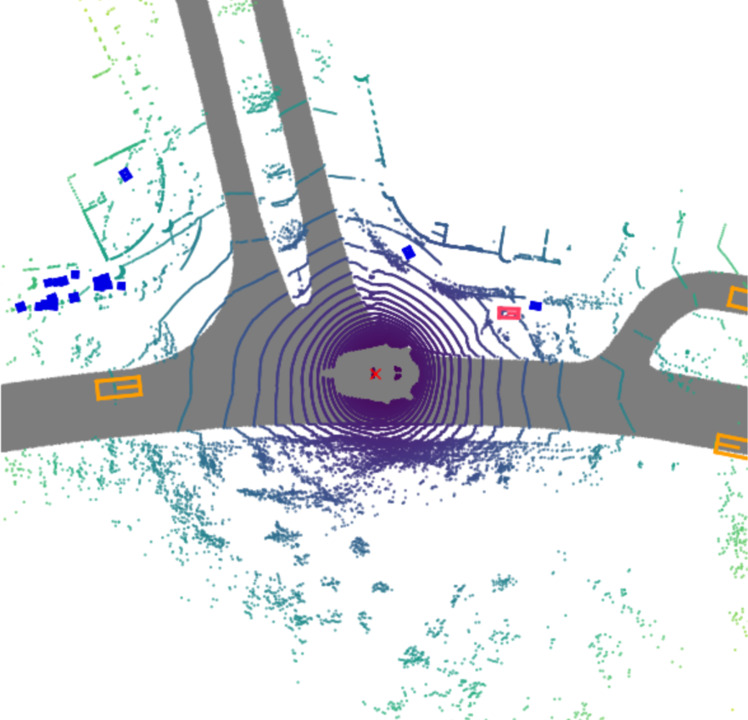}
\end{minipage}
}\hspace{0.27cm}
{
\begin{minipage}{0.12\textwidth} 
\centering
\includegraphics[height=2.5cm,width=2.5cm]{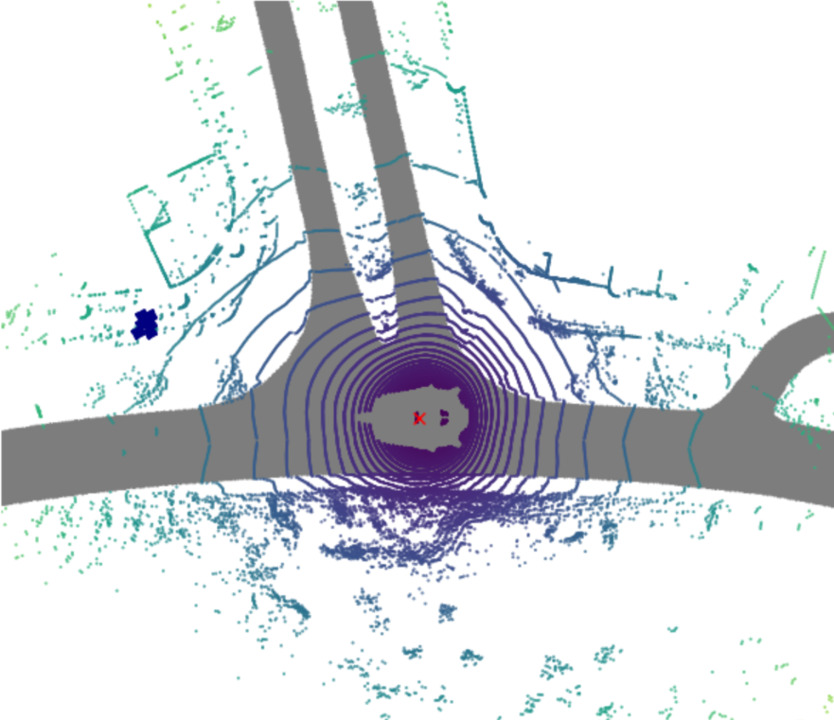}
\end{minipage}
}\hspace{0.27cm}
{
\begin{minipage}{0.12\textwidth} 
\centering
\includegraphics[height=2.5cm,width=2.5cm]{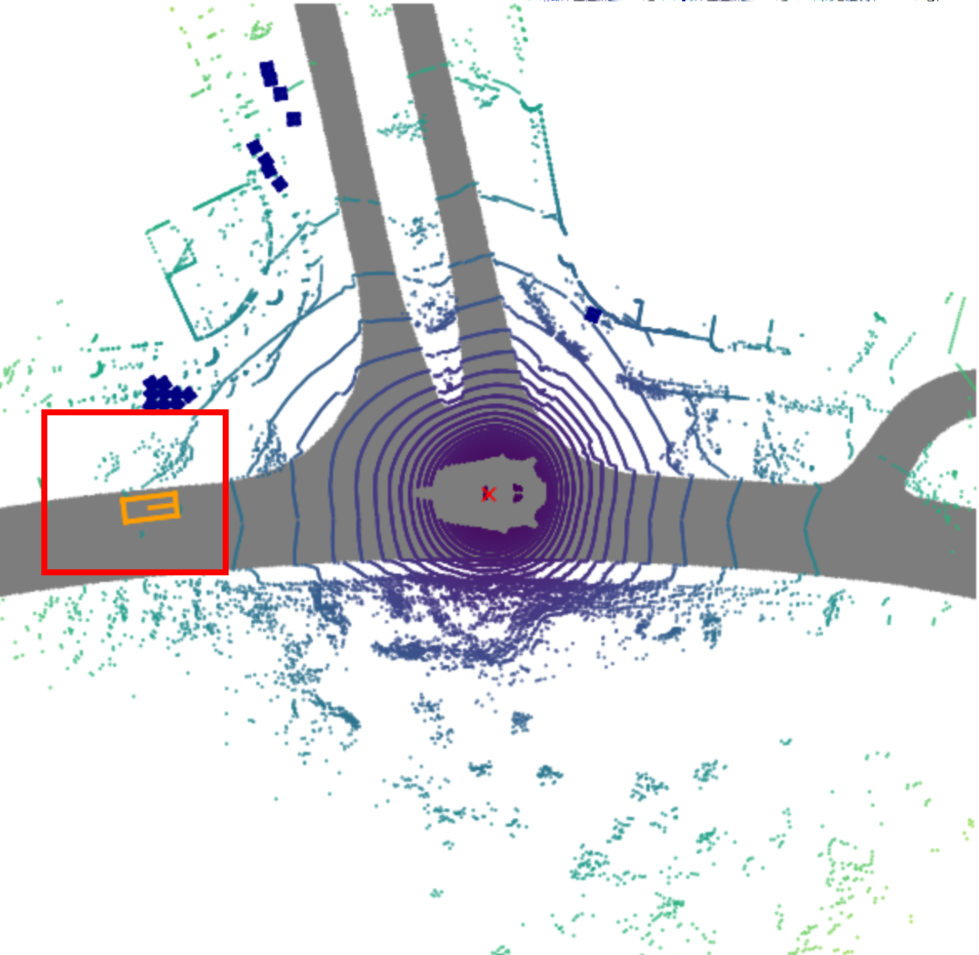}
\end{minipage}
}

\hspace{-1cm}
{
\begin{minipage}{0.01\textwidth} 
\centering
\rotatebox{90}{\footnotesize{$128\times128$}}
\end{minipage}
}\hspace{-0.1cm}
{
\begin{minipage}{0.12\textwidth} 
\centering
\includegraphics[height=2.5cm,width=2.5cm]{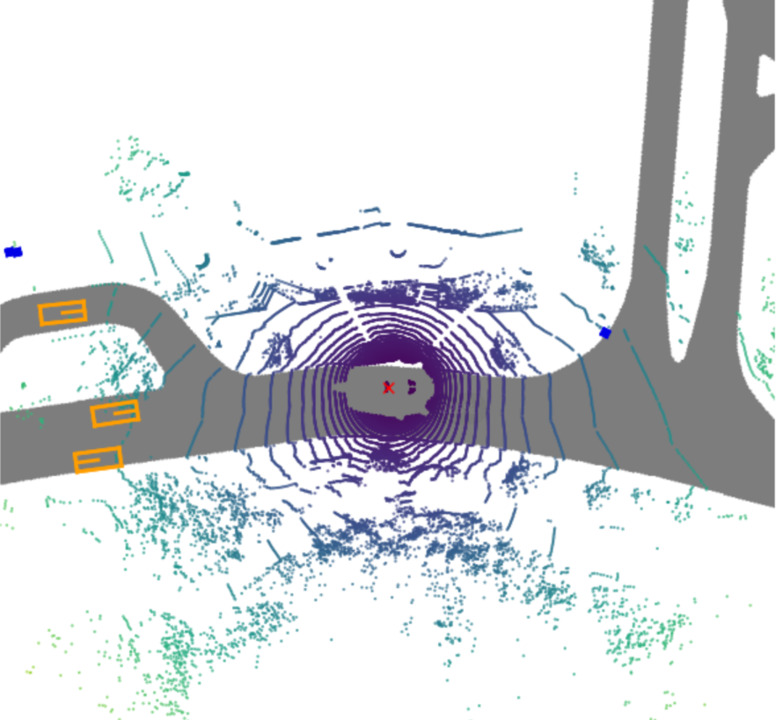}
\end{minipage}
}\hspace{0.27cm}
{
\begin{minipage}{0.12\textwidth} 
\centering
\includegraphics[height=2.5cm,width=2.5cm]{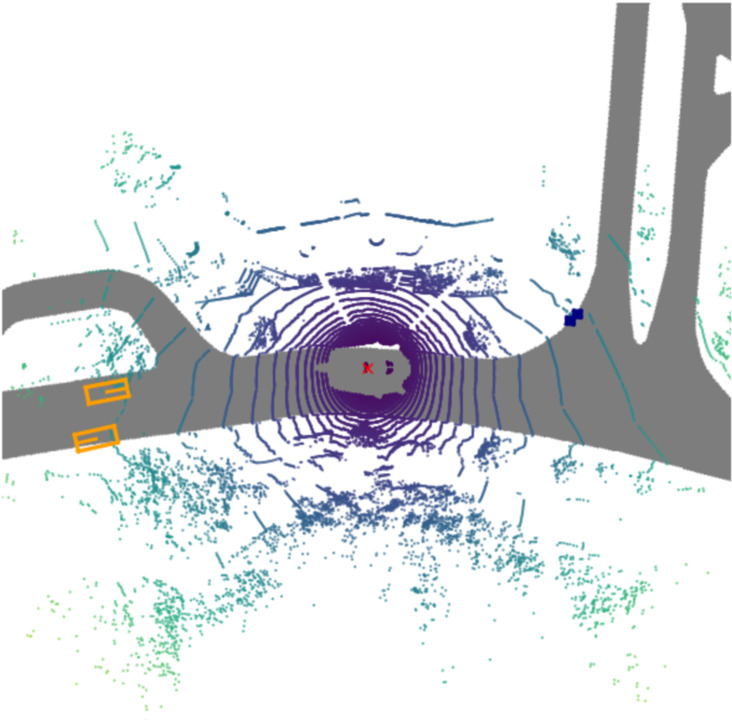}
\end{minipage}
}\hspace{0.27cm}
{
\begin{minipage}{0.12\textwidth} 
\centering
\includegraphics[height=2.5cm,width=2.5cm]{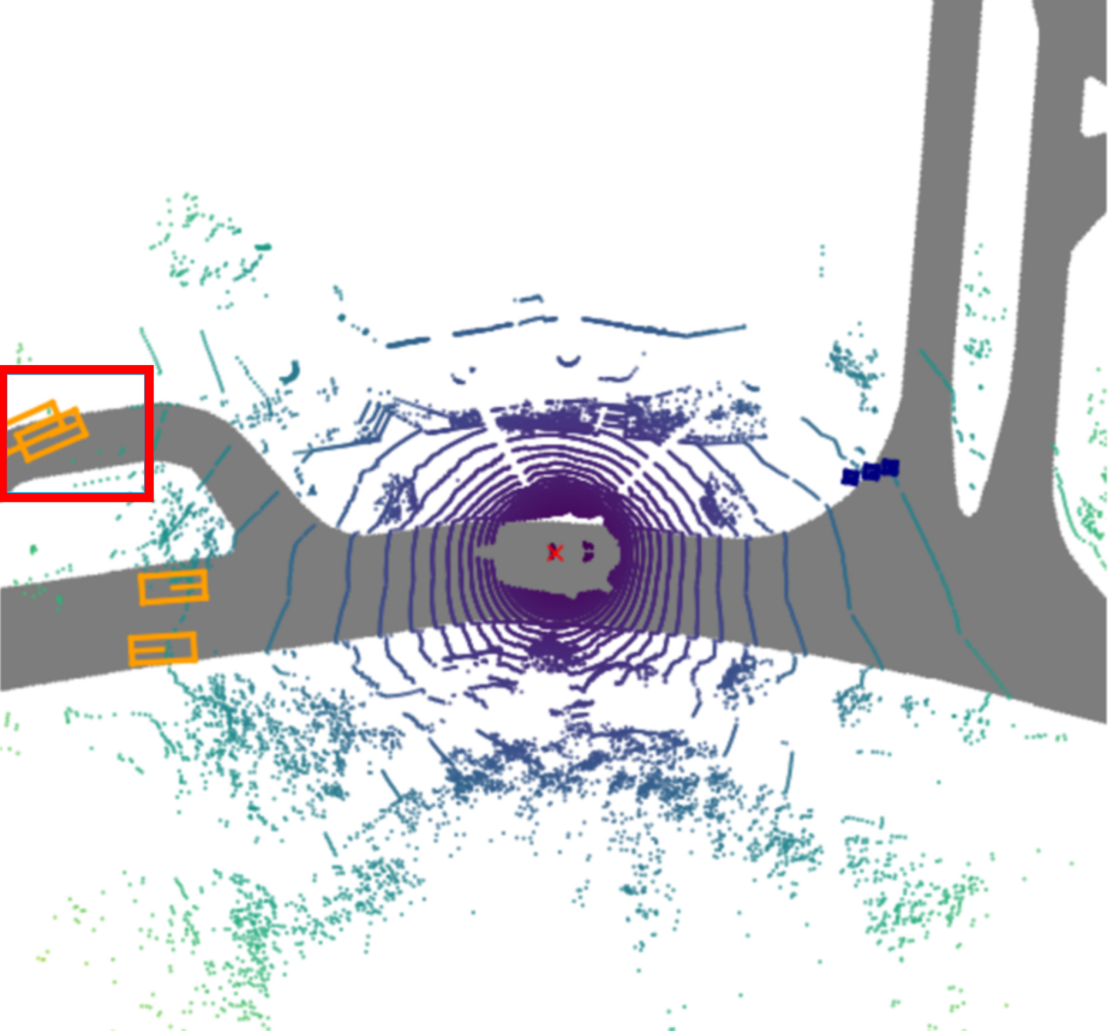}
\end{minipage}
}

\hspace{-1cm}
{
\begin{minipage}{0.01\textwidth} 
\centering
\rotatebox{90}{\footnotesize{$200\times200$}}
\end{minipage}
}\hspace{-0.1cm}
{
\begin{minipage}{0.12\textwidth} 
\centering
\includegraphics[height=2.5cm,width=2.5cm]{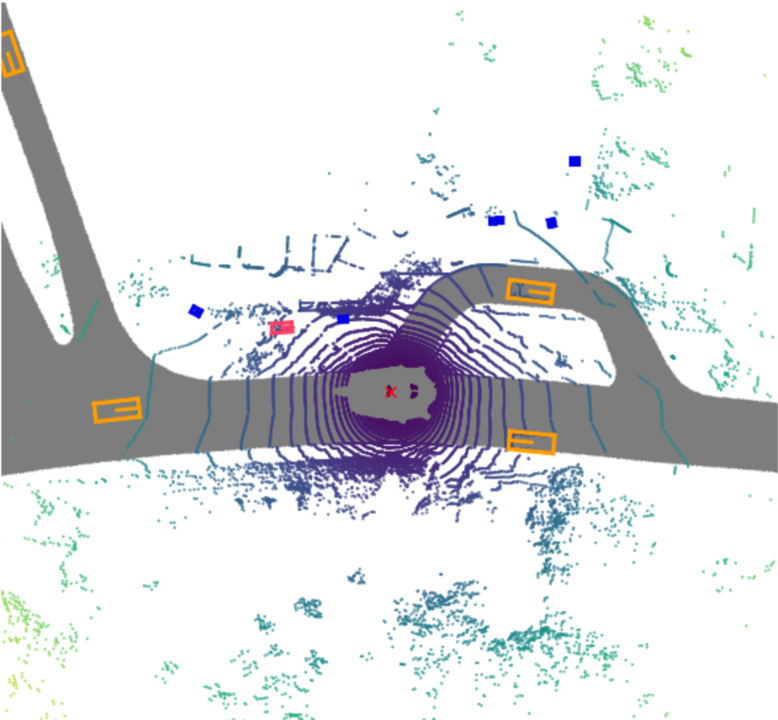}
\end{minipage}
}\hspace{0.27cm}
{
\begin{minipage}{0.12\textwidth} 
\centering
\includegraphics[height=2.5cm,width=2.5cm]{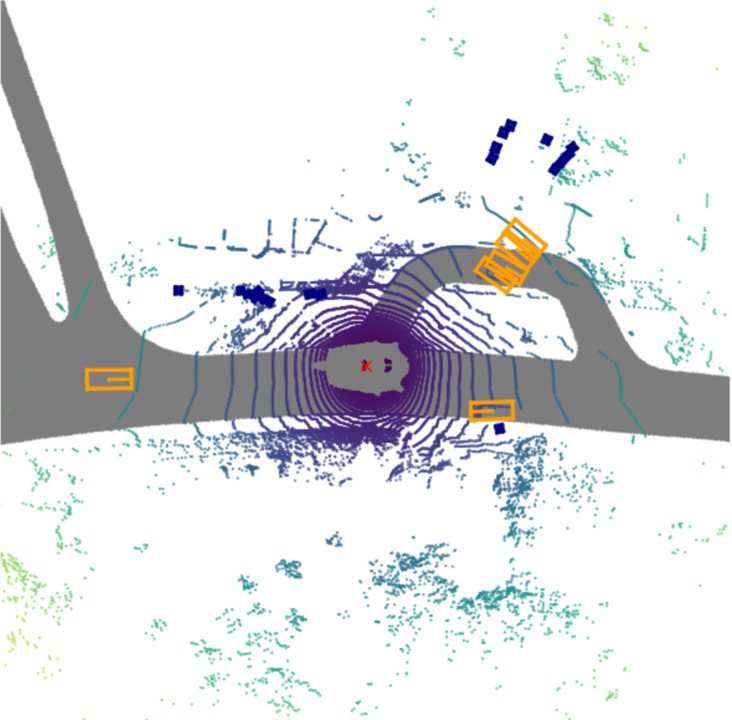}
\end{minipage}
}\hspace{0.27cm}
{
\begin{minipage}{0.12\textwidth} 
\centering
\includegraphics[height=2.5cm,width=2.5cm]{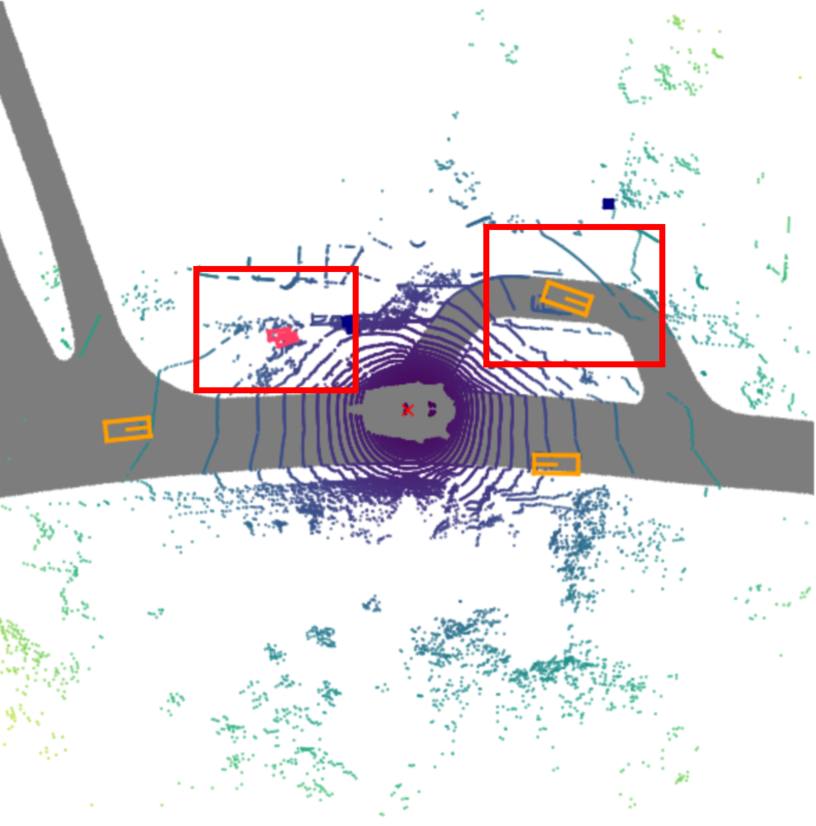}
\end{minipage}
}

\hspace{-0.85cm}
{
\begin{minipage}{0.01\textwidth} 
\centering
\rotatebox{90}{\footnotesize{$256\times256$}}
\end{minipage}
}\hspace{-0.23cm}
\subfigure[Ground-truth]
{
{
\begin{minipage}{0.12\textwidth} 
\centering
\includegraphics[height=2.5cm,width=2.5cm]{Figures/GT-256.jpg}
\end{minipage}
}
}
\subfigure[BEVDet]
{
{
\begin{minipage}{0.12\textwidth} 
\centering
\includegraphics[height=2.5cm,width=2.5cm]{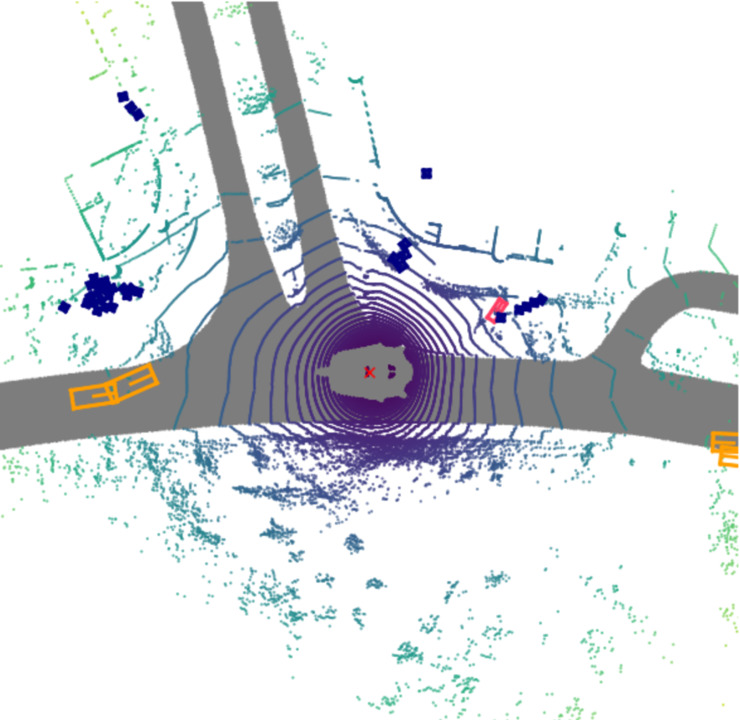}
\end{minipage}
}
}
\subfigure[Ours]
{
{
\begin{minipage}{0.12\textwidth} 
\centering
\includegraphics[height=2.5cm,width=2.5cm]{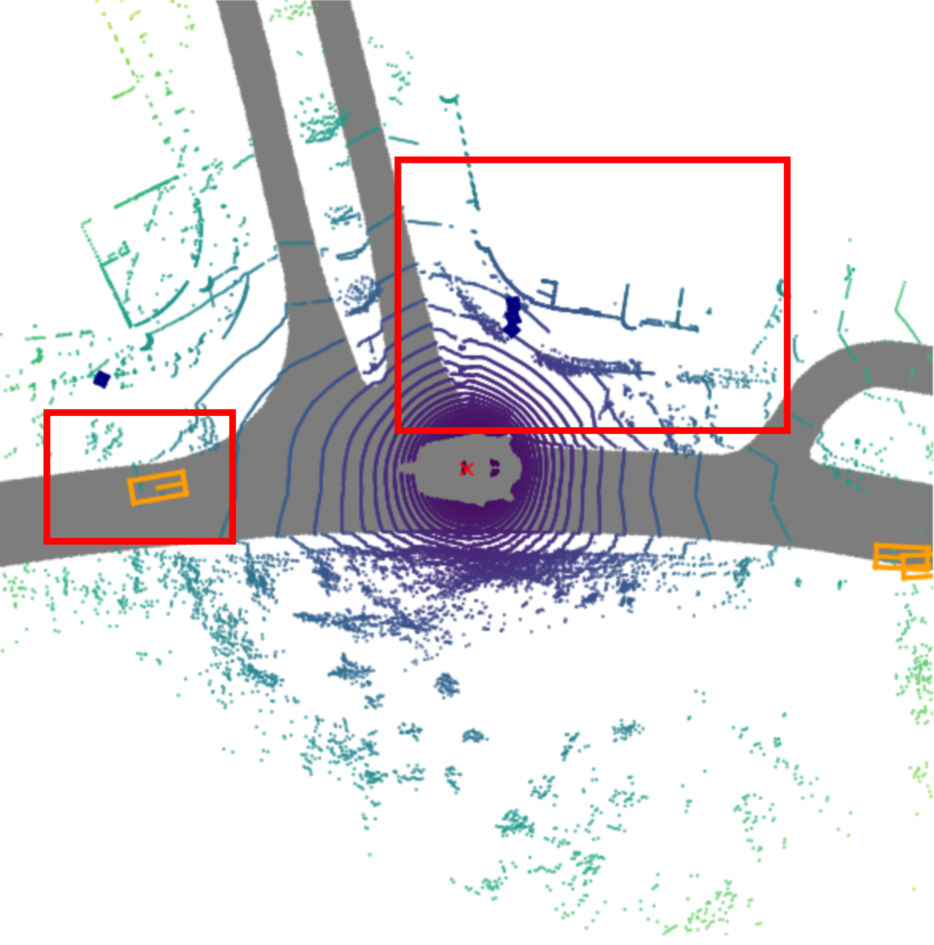}
\end{minipage}
}
}

\caption{Visualization of the experimental results in Table.~\ref{tab:seen}~\ref{tab:unseen}. 
We highlight the area with more accurate predictions for our method with blue rectangles.}
\label{Fig:visulization}
\end{figure}

\begin{table*}[htbp]
  \centering
  \caption{Quantitative evaluation of recent vision-centric 3D detection works on the nuScenes val set. Among all methods, DETR3D, PETR, BEVDet and our method are trained with CBGS~\cite{cbgs}.}
    \scalebox{0.79}{\begin{tabular}{c|c|c|c|c|c|c|c|c|c}
    \hline
    Method & Backbone & Size  & NDS$\uparrow$   & mAP$\uparrow$   & mATE$\downarrow$  & mASE$\downarrow$  & mAOE$\downarrow$  & mAVE$\downarrow$  & mAAE$\downarrow$ \\
    \hline
    CenterNet~\cite{centernet} & DLA   &   --    & 0.328 & 0.306 & 0.716 & 0.264 & 0.609 & 1.426 & 0.658 \\
    \hline
    FCOS3D~\cite{FCOS3D} & ResNet-50 & 1600×900 & 0.368 & 0.288 & 0.777 & 0.266 & 0.544 & 1.228 & 0.170 \\
    \hline
    DETR3D~\cite{DETR3D} & ResNet-50 & 1600×900 & 0.373 & 0.302 & 0.811 & 0.282 & 0.493 & 0.979 & 0.212 \\
    \hline
    PGD~\cite{PGD}   & ResNet-50 & 1600×900 & 0.394 & 0.320  & 0.735 & 0.266 & 0.492 & 1.114 & 0.170 \\
    \hline
    Ego3RT~\cite{ego3rt} & ResNet-50 & 1600×900 & 0.380  & 0.332 & 0.706 & 0.281 & 0.663 & 0.964 & 0.249 \\
    \hline
    PETR~\cite{petr}  & ResNet-50 & 1056×384 & 0.381 & 0.313 & 0.768 & 0.278 & 0.564 & 0.923 & 0.225 \\
    \hline
    BEVDet~\cite{BEVDet} & ResNet-50 & 704×256 & 0.379 & 0.298 & 0.725 & 0.279 & 0.589 & 0.860  & 0.245 \\
    \hline
    BEVDet~\cite{BEVDet} & ResNet-101 & 704×256 & 0.381 & 0.302 & 0.722 & 0.269 & 0.543 & 0.900   & 0.269 \\
    \hline
    BEVDet~\cite{BEVDet} & Swin-tiny & 704×256 & 0.392 & 0.312 & 0.691 & 0.272 & 0.523 & 0.909 & 0.247 \\
    \hline
    PersDet~\cite{PersDet} & ResNet-50 & 704×256 & 0.389 & 0.319 & 0.676 & 0.284 & 0.589 & 0.924 & 0.229 \\
    \hline
    CaDDN~\cite{CADDN} & ResNet-50 & 704×256 & 0.370  & 0.294 & 0.702 & 0.283 & 0.579 & 0.988 & 0.222 \\
    \hline
    Ours  & Swin-tiny & 704×256 & \textbf{0.398} & \textbf{0.321} & \textbf{0.669} & \textbf{0.275} & \textbf{0.494} & \textbf{0.956} & \textbf{0.231} \\
    \hline
    \end{tabular}}%
  \label{tab:Quantitative}%
\end{table*}%

\subsection{Implementation Details}
\label{subsec:imp}
For a fair comparison, We employ the pretrained Swin-Transformer as our 2D backbone network, following existing works~\cite{BEVDepth,bevformer,beverse}. Our perception range is $[-51.2m,51.2m]$ along the X and Y axis.
For the view symmetry of multi-view cameras, the radius and azimuth are set to $(144,360)$, which represents the polar perception range.
The input multi-view image is rescaled to (256, 704) for prototype verification.  The network is trained with CBGS~\cite{cbgs} on 4 RTX 3090 GPUs with total batch size of 8. During the inference, we apply our model to different BEV grid resolutions with the fixed perception range, where we sample the polar feature maps to rectangle Cartesian BEV feature maps with various grid resolutions for adapting to different computational budgets. For the inference speed, our method is 9 FPS for $64\times64$ BEV resolution and 6.5 FPS for $256\times256$ BEV resolution.

\subsection{One Training for Multiple Deployments}
\label{subsec:imp1}

In this section, we show the results of our method in the one-training-multiple-deployments setting. Specifically, the proposed method is trained with only one head (128$\times$128 feature map) while it would conduct the inference on multiple feature maps with different size, like 200 $\times$ 200 and 256 $\times$ 256, which also denotes the different computational budgets. Due to the resolution-invariant property of Polar feature space, our method can adapt to different BEV grid sizes by various degrees of sampling. While for these existing methods, including BEVDet~\cite{BEVDet} and BEVerse~\cite{beverse}, they cannot be employed to other resolution directly. To make a comparison, a modification of these methods is applied where we perform the interpolation over the BEV feature map and keep the detection head untouched to adapt them to different output sizes. 
For some other related BEV perception works~\cite{DETR3D,petr,petrv2}, they do not build the explicit BEV feature map and lack the ability for diverse deployments with various computing budgets, so we do not compare with them here.

As shown in Table~\ref{tab:seen}, all methods are trained with feature map size of 128$\times$128 and different feature maps are used for inference (200$\times$200 and 256 $\times$256). Note that different sizes of output features indicate the different resolution of BEV grid, also indicate the different computational budgets. It can be found that our method achieves the consistent performance gain compared to these baseline methods in almost all metrics. Specifically, when we perform the inference on different output feature maps~(200$\times$200 and 256 $\times$256), our method shows decent generalization capability with marginal performance drop, like -2\% (128$\times$128 $\longrightarrow$ 200$\times$200) and -5\% (128$\times$128 $\longrightarrow$ 256$\times$256) in terms of NDS, while BEVDet~\cite{BEVDet} gives a large degradation, -7\% and -12\%. 
{Note that BEVerse performs better in mAVE, mAAE and NDS because the multi-frame images are leveraged as input. In our work, we focus more on the generalization ability among different output resolutions, which is clearly demonstrated from the marginal mAP drops when inferencing with an output different resolution.}

We also provide visualization results in Figure.~\ref{Fig:visulization}. From the highlighted areas, it is obvious to see that our method has more accurate detection ability, especially when adapting to other feature map scales. It further demonstrates the superiority of our method.

\noindent{\textbf{More challenging deployments}}~~~ In Table~\ref{tab:unseen}, we conduct the inference on a more challenging and flexible setting, where 64$\times$64 and 96 $\times$96 output feature maps are conducted and the training head is also 128 $\times$ 128. It is noted that these output feature maps are flexible and unseen during training. From the results, we can find that our method achieves the best generalization performance with the least performance drop compared to baseline methods, indicating its good generalization capability in the challenging setting.  



\subsection{Ablation Study and Qualitative Evaluation}
To verify the effectiveness of important modules of our network, we conduct ablation study and the results are shown in Table.~\ref{tab:ablation}. We can see that the polar-based BEV feature representation is superior to pure Cartesian feature representation and the multi-scale feature interaction process further improve the performance. We also show the quantitative evaluation results for vision-centric BEV perception in Table.~\ref{tab:Quantitative}. Compared with other methods, our method is decent and competitive.

\section{CONCLUSION}
In this paper, we propose a novel vision-centric BEV perception approach for autonomous driving, which can be flexibly deployed to multiple platforms with various computing budgets. In particular, we leverage the Polar-represented BEV feature map to solve this problem based on its good property that the feature along rays can be easily mapped to Cartesian grid-represented feature with arbitrary resolutions. We also design an effective interaction module to further enhance the feature representation by multi-scale feature interactions. Extensive experiments show that our method has superior generalization capability for inferring novel feature map scales with only one training, which is significant for the real deployment in diverse scenarios.

\section{Acknowledgements}

This work was supported by NSFC (No.62206173), Shanghai Sailing Program (No.22YF1428700), and Shanghai Frontiers Science Center of Human-centered Artificial Intelligence (ShangHAI).

\printbibliography

\end{document}